\documentclass[letterpaper, 10 pt, conference]{ieeeconf}  

\IEEEoverridecommandlockouts                              

\usepackage{bm}
\usepackage{algorithm}
\usepackage{algorithmicx}
\usepackage{graphicx}
\usepackage{amsmath}
\usepackage{amssymb}
\usepackage{algpseudocode}
\usepackage{subfig}
\graphicspath{{figures/}}

\title{\LARGE \bf
Submap-based Pose-graph Visual SLAM:\\ A Robust Visual Exploration and Localization System$^{*}$
\footnoterule\thanks{$^{*}$ The work in this paper is supported by the National Natural Science Foundation of China (61603103, 61673125), the Natural Science Foundation of Guangdong of China (2016A030310293), and the Major Scientific and Technological Special Project of Guangdong of China (2016B090910003).}}

\author{
\authorblockN{Weinan Chen\authorrefmark{2}, Lei Zhu\authorrefmark{2}, Yisheng Guan\authorrefmark{2}, C. Ronald Kube\authorrefmark{3}, Hong Zhang*\authorrefmark{3}} \vspace{2mm}
\authorblockA{\authorrefmark{2}Biomimetic and Intelligent Robotics Lab, Guangdong University of Technology, Guangzhou, China}
\authorblockA{\authorrefmark{3}Department of Computing Science, University of Alberta, Edmonton, Canada}}

\begin{document}

\maketitle
\thispagestyle{empty}
\pagestyle{empty}

\begin{abstract}

For VSLAM (Visual Simultaneous Localization and Mapping), localization is a challenging task, especially for some challenging situations: textureless frames, motion blur, etc.. To build a robust exploration and localization system in a given space or environment, a submap-based VSLAM system is proposed in this paper. Our system uses a submap back-end and a visual front-end. The main advantage of our system is its robustness with respect to tracking failure, a common problem in current VSLAM algorithms. The robustness of our system is compared with the state-of-the-art in terms of average tracking percentage. The precision of our system is also evaluated in terms of ATE (absolute trajectory error) RMSE (root mean square error) comparing the state-of-the-art. The ability of our system in solving the ``kidnapped'' problem is demonstrated. Our system can improve the robustness of visual localization in challenging situations.

\end{abstract}

\section{Introduction}
\label{section_introduction}

Many robot applications require a mobile robot to explore an unknown environment and construct its model. Environment exploration with vision is popularly studied in robotics via VSLAM. However, the explored space in this study is given and limited. In many scenarios, we are able to explore a certain space repeatedly, especially for the indoor applications. We refer to such an exploration as ``dense exploration'' in this paper.

For the keyframe-based VSLAM systems, reconstruction of the environment model with known poses is not difficult, and neither is the problem of localization in a known map. However, resolving both the environment model and localization of the robot at the same time is extremely difficult, and has been a focus of robotics research for over 20 years. In the case of indoor VSLAM, there are a number of challenges such as textureless walls, motion blurred frames and illumination changes \cite{c1}. Even given enough time, the existing VSLAM systems \cite{c2}\cite{c3}\cite{c4} do not work all the time, since they rely on a back-end with a single graph and are unable to maintain working continuously. In contrast, by using multiple subgraphs, it is much easier for the robot to maintain tracking in one of the local subgraphs than in a global graph.

Instead of a single graph, a multiple submap graph is used for robustness improvement in this study. Submap-based graph is an approach that represents a complete environment using multiple subgraphs, which has been studied for some time \cite{c5}\cite{c6}\cite{c7}. Also, the merging of submaps is a key problem in submap-based systems, and many related studies have been proposed \cite{c8}\cite{c9}\cite{c10}. As a VSLAM system, we try to merge the submaps using place recognition \cite{c11}\cite{c12} in the front-end and a robust optimization \cite{c13}\cite{c14} in the back-end.

A submap-based VSLAM system is introduced in this paper. A pose-graph containing multiple submaps is built and maintained. To work with such a back-end, a multi-constraint front-end is designed and introduced. As evaluated by the experiments, our system can improve the robustness and has a better precision than the state-of-the-art. In addition, the ability of solving the ``kidnapped'' problem with the map built by our system is demonstrated.

The paper is organized as follows. Section \ref{section_system_overview} presents the overview of the whole system, including the framework and the basic idea. Section \ref{section_front_end} details the front-end, including the tracking and place recognition. Section \ref{section_back_end}, a submap-based back-end is proposed. Section \ref{section_implementation} introduces other details and the implementation of our system. Section \ref{section_experiments} provides an evaluation and discussion of our system. Section \ref{section_conclusion} concludes and summaries the paper.

\section{System Overview}
\label{section_system_overview}

\begin{figure}[htp]
\centering
\includegraphics[width=0.3\textwidth]{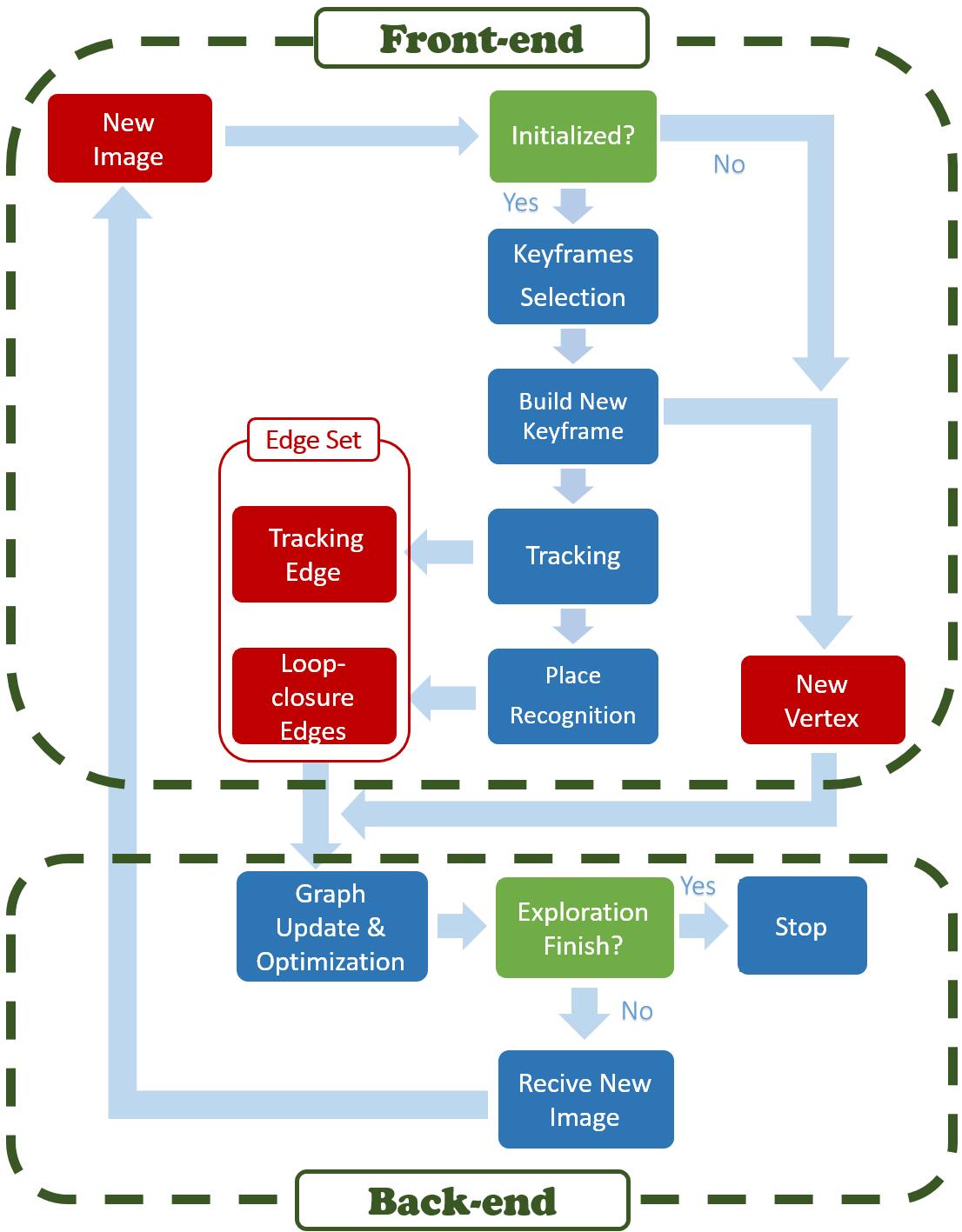}
\caption{Framework of our system}
\label{fig_framework}
\end{figure}

The framework of our system is shown in Fig. \ref{fig_framework}, where the red boxes are the inputs/outputs and the green boxes are the decisions (the same representations will be applied to all the flow charts in this paper).

As a VSLAM system, it can be divided into two parts: the front-end and the back-end. To get an optimized performance of the dense exploration, a multi-constraint front-end and a submap-based back-end is proposed to improve the robustness and reliability.

The basic idea of the front-end is: for a new keyframe, we try to build an edge set using place recognition and tracking, and update the pose-graph for optimization. After receiving a new frame, if the system has been initialized (the last keyframe is not empty), we select the keyframe and store it; if not, we take the frame as the first keyframe directly.

For a keyframe, we try to build two types of constraints: ``tracking edges'' from the last keyframe and ``loop-closure edges'' from the place recognition module. With these constraints, an optimization can be found to calculate the pose of the new keyframe. To calculate the constraint, a feature-based method is applied to find the relative pose between two frames, and a visual BoW (Bag-of-Word) approach is used for place recognition. The details of the front-end will be introduced in Section \ref{section_front_end}.

As for the back-end, a pose-graph is built and maintained. Once no edge can be established for a new keyframe, a new submap is built and we try to maintain tracking in that submap. Such a mechanism keeps the system ``alive'' even after tracking lost happens. We merge the submaps by using the loop-closure edges, and the switchable factors are added to those loop-closure edges to improve the robustness of submap merging. In Section \ref{section_back_end}, the back-end will be detailed. The illustration of submap maintaining and optimization is shown in Fig. \ref{fig_back_end}.

In addition, an exploration end-condition is proposed to prevent infinite exploration, which takes the density of vertices into consideration. By the multi-constraint front-end and submap-based back-end, the robustness of visual exploration is improved, and the robust visual localization is achieved by the map built with our proposed end-condition.

\section{Multi-constraint Front-end}
\label{section_front_end}

\begin{figure}[htp]
\centering
\includegraphics[width=0.4\textwidth]{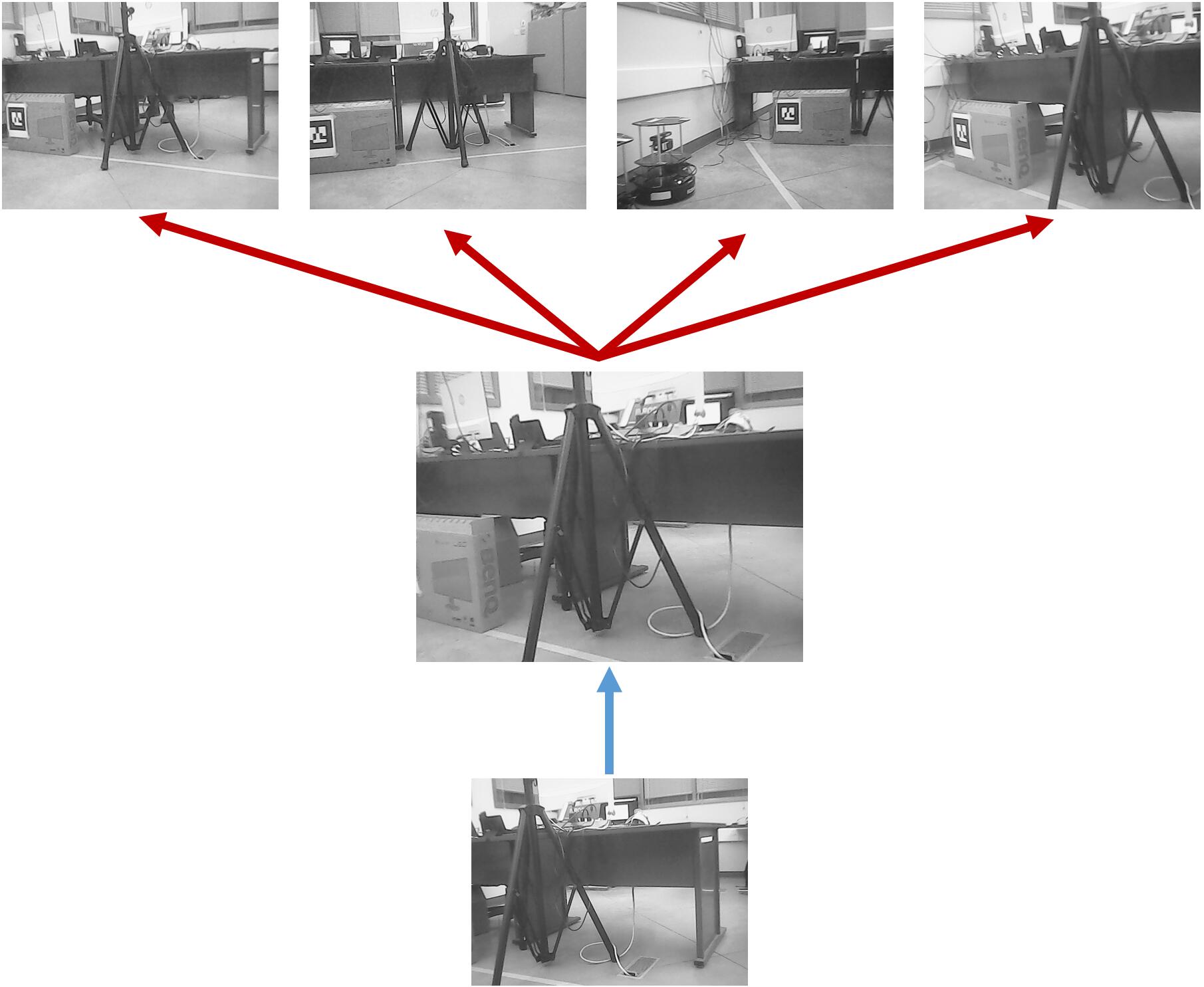}
\caption{An example of multi-constraint front-end, where the blue arrow is the tracking edge from the last keyframe to the current frame, and the red arrows are the loop-closure edges connected to the place recognition results.}
\label{fig_multi_constraint}
\end{figure}

In the front-end, a graph is built with camera poses, tracking edges and loop-closure edges by VO (visual odometry). To calculate the constraints between the frames, a straight feature-based tracking method is designed, and a strict condition is performed to reject the low quality edges. As for place recognition, an image global descriptor is extracted and applied for image matching. For submap merging, a lax condition is executed to add more loop-closure edges to the graph. An example is shown in Fig. \ref{fig_multi_constraint}. By considering more keyframes for localization in Multi-constraint Front-end, the robustness can be improved.

\subsection{Keyframe selection}

As a keyframe-based VSLAM system, we select the keyframes from the image sequence. There are three conditions for inserting keyframes:

(1) The normal of the relative transformation between the current frame and the last keyframe should be larger than a given threshold, which means the current image is different from the last keyframe spatially.

(2) The distance between the global descriptor of the current frame and the last keyframe should be larger than a given value to reject the frames that are too similar to the last keyframe in appearance.

(3) When neither tracking edge nor loop-closure edge can be established for a frame, the frame is ``lost'', this frame is treated as the first keyframe of a new submap.

The role of the first condition is similar to that of the second condition. However, considering the normalization of the translation in the relative pose calculation, the image similarity (represented by the distance of global descriptor) is considered in the second condition.

To maintain the possibility of a dense reconstruction after exploration, and meet the requirement of global descriptor extraction, a complete image is saved in each keyframe. Some common information of a keyframe (keypoints, index, pose, etc.) and an edge set is built to store all the constraints.

\subsection{Tracking}

The tracking edge is the relative pose between the current frame and the last keyframe. We calculate the relative pose through a two-frames feature-based pipeline: feature detection, feature matching, fundamental matrix finding and essential matrix decomposing. In addition, a verification is designed to reject a low quality result.

\begin{figure}[ht]
\centering
\includegraphics[width=0.35\textwidth]{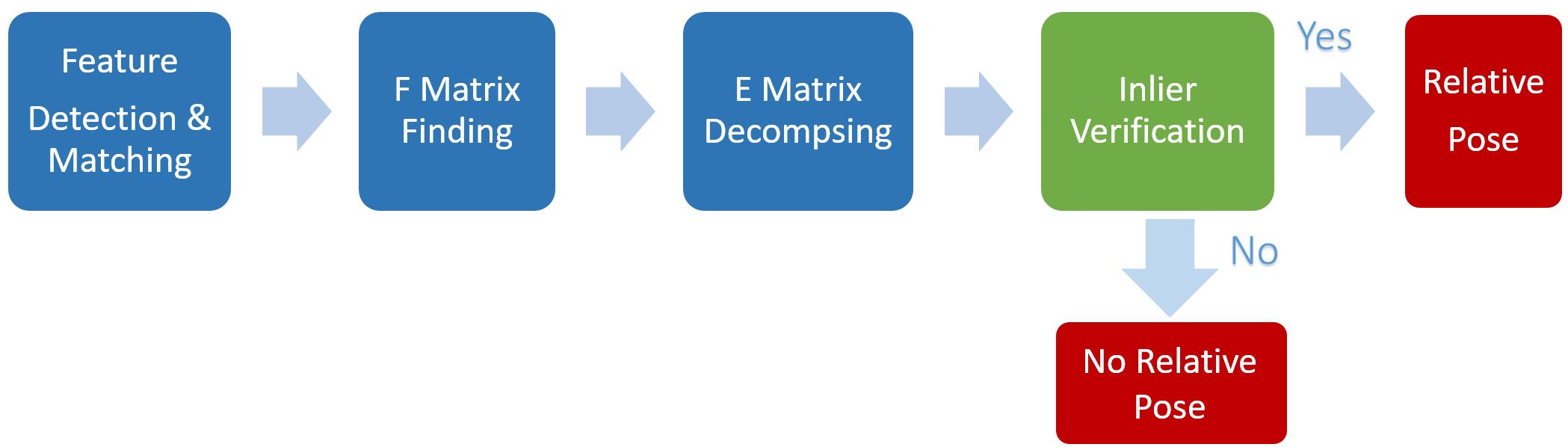}
\caption{Pipeline of the constraint building between two frames}
\label{fig_front_end}
\end{figure}

A feature detection method similar to \cite{c15} is used, a grid is place on an image during feature detection, for an even feature distribution. The FAST detector and ORB descriptor are selected for their efficiency. Also, RANSAC \cite{c23} is applied to fundamental matrix finding and essential matrix decomposing, and the size of inlier features $s$ is recorded. The pipeline is illustrated in Fig. \ref{fig_front_end}.

To improve the precision of tracking edges added to the graph, that is solved by calculating a homography matrix and a fundamental matrix simultaneously as in \cite{c15}, we set a strict condition with respect to $s$ to accept the high quality result only. If $s$ is large enough, the constraint is accepted; otherwise, the constraint is rejected and no tracking edge is established. However, thanks to the design of multi-constraint front-end and submap-based back-end, the system still updates the graph even if no tracking constraint can be built.

\subsection{Place Recognition}

Loop-closure edges are established by place recognition, which finds the places that have been visited before. A global descriptor of each frame is extracted by the DBoW2 \cite{c12} descriptor. To decrease the computation time and increase the frame rate, the place recognition results are simply obtained by ranking the distance of global descriptors between the current frame and all the existing keyframes. Only the keyframes that are ``far enough from the current frame'' are considered, which means a distance between the index of current frame and the place recognition results is required.

After global descriptor matching, we pick the first $k$ matching results as the potential loop closures, and all of them are sent to loop closure verification before inserting the constraints as a set of new edges for the new keyframe. For the loop closure verification, we do the same thing as is mentioned in the last subsection: matching the features and calculating $s$ to verify the loop closure candidates. However, a lax condition with respect to $s$ is executed for adding more loop-closure edges to the graph to increase the possibility of submap merging, instead of a strict condition while establishing tracking edges.

As is known, the relative pose between two the monocular visual observations has an unknown scale. Therefore, we store more than one matching result of place recognition and build multiple constraints for a new frame. By least-square optimization, the estimation error caused by the different unknown scales can be iteratively minimized. This idea will be explained with a ``spring model'' in the Section \ref{section_back_end}.

\section{Submap-based Back-end}
\label{section_back_end}

After keyframe insertion and constraint calculation, a pose-graph is built. To overcome the tracking failure and get a robust performance, a submap-based graph is built.

\begin{figure}[htp]
\centering
\includegraphics[width=0.25\textwidth]{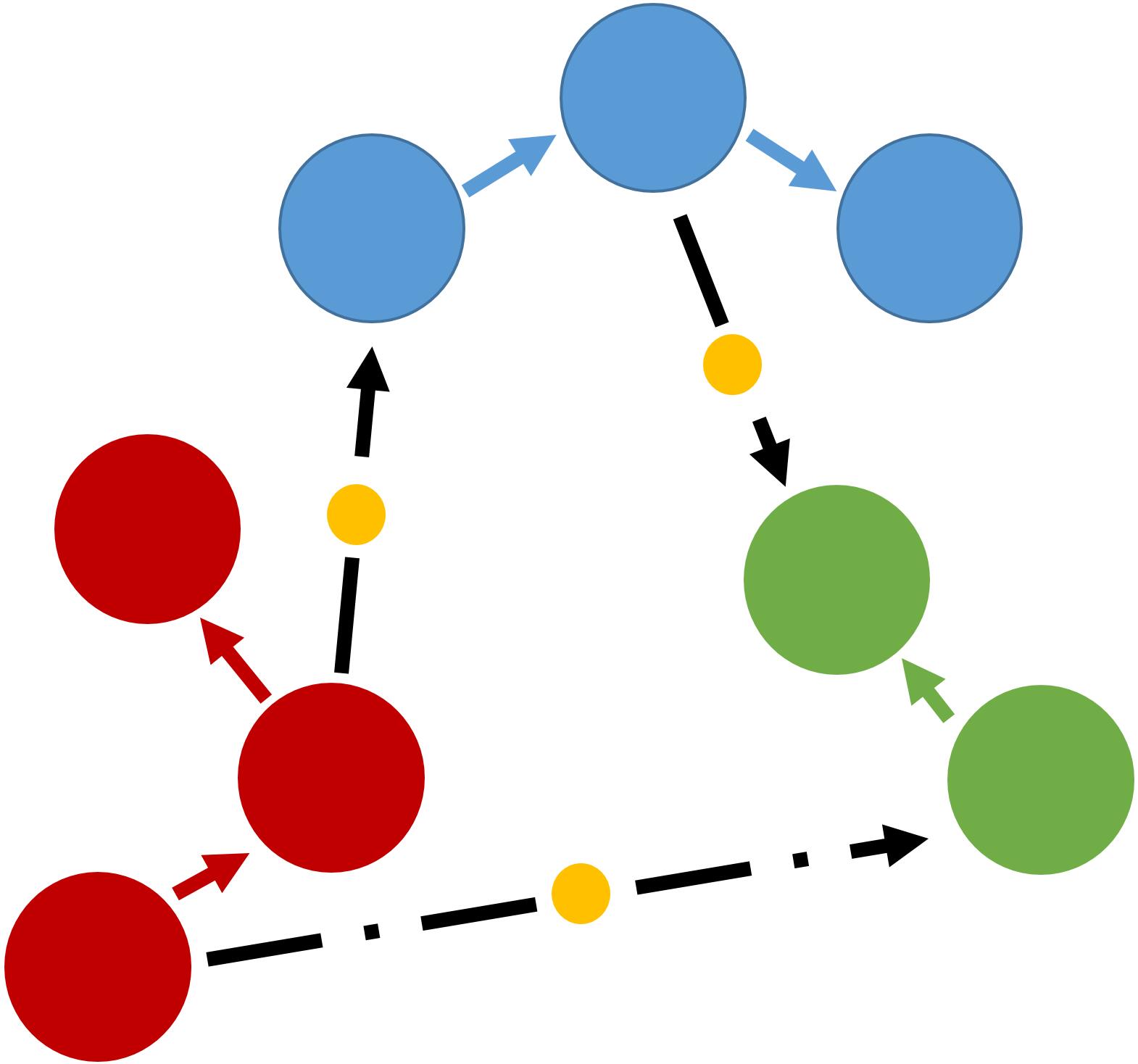}
\caption{The basic idea of submap-based graph. Three submaps (red nodes, blue nodes and green nodes) are built within a global graph, the solid arrows are the tracking edges within a submap, the black dotted arrows are the loop-closure edges for submap merging, and the switchable factors (yellow nodes) are added to all the loop-closure edges.}
\label{fig_submap}
\end{figure}

We optimize the vertex poses and merge the submaps using the tracking edges and loop-closure edges with a weighted information matrix, and the switchable factors close the outlier loop-closure edges during optimization. The basic idea of submap-based graph is illustrated in Fig. \ref{fig_submap}.

\subsection{Definition}

The vertex representing a camera pose is defined as a $SE(3)$ matrix, and the switchable factor added to the loop-closure edges is represented as a float value. As for the constraints between the vertices, two types of constraints are established and optimized: tracking edges and loop-closure edges. The edge is also represented as a $SE(3)$ matrix.

\begin{figure}[htp]
\centering
\includegraphics[width=0.4\textwidth]{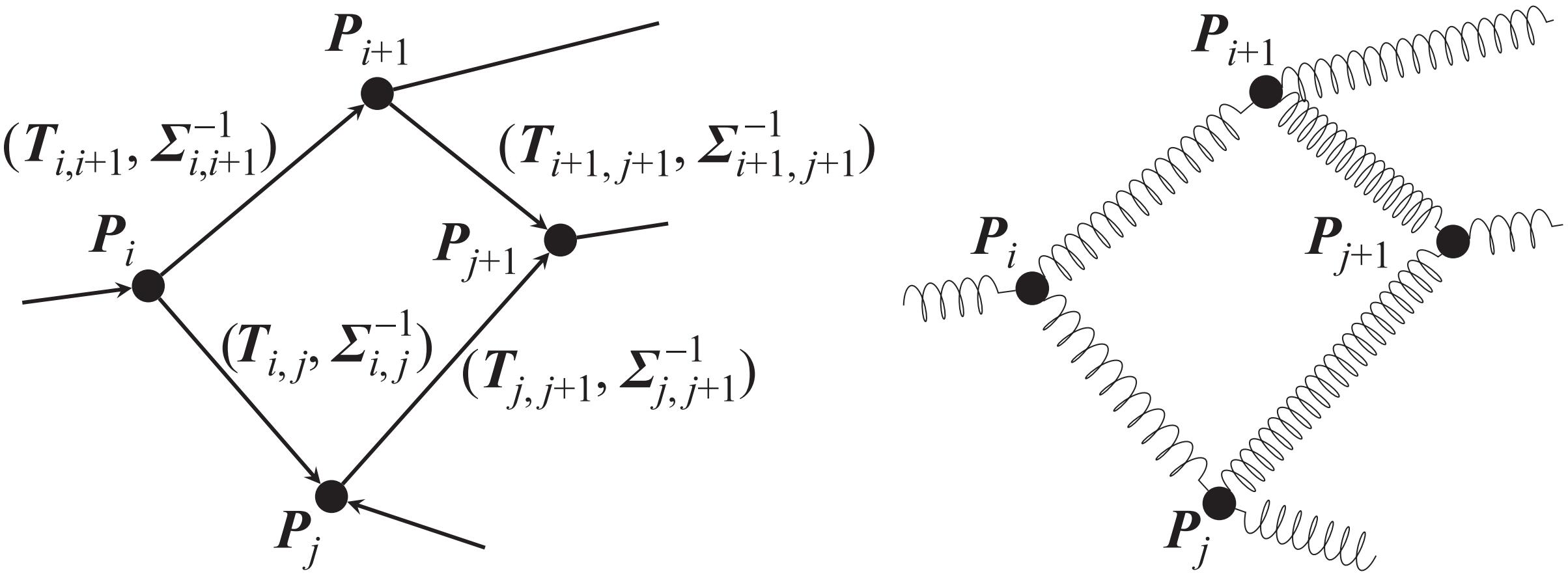}
\caption{Spring model \cite{c17}}
\label{fig_spring model}
\end{figure}

For optimization, a least-square optimization is performed to minimize the error of all the edges, which means we optimize the whole graph every time instead of a sliding window optimization. Even though global optimization is time-consuming, it is required for submap merging. For decreasing the error of the edges with an unknown scale, we build multiple non-scale edges for a vertex. The graph is represented as a ``spring model \cite{c16}'', as is illustrated in Fig. \ref{fig_spring model}.

Even the edge has an unknown scale in terms of the translation, it gives a constraint in the heading of the translation, which can be treated as a direction of the force of springs. If the vertex has more than one edge, it can be considered to be pulled by many springs from different directions, and the pose of the vertex can be obtained when the spring forces are balanced after optimization. Also, the scale of all the edges will be aligned after optimization.

\subsection{Approach}

\begin{figure}[htp]
\centering
\includegraphics[width=0.35\textwidth]{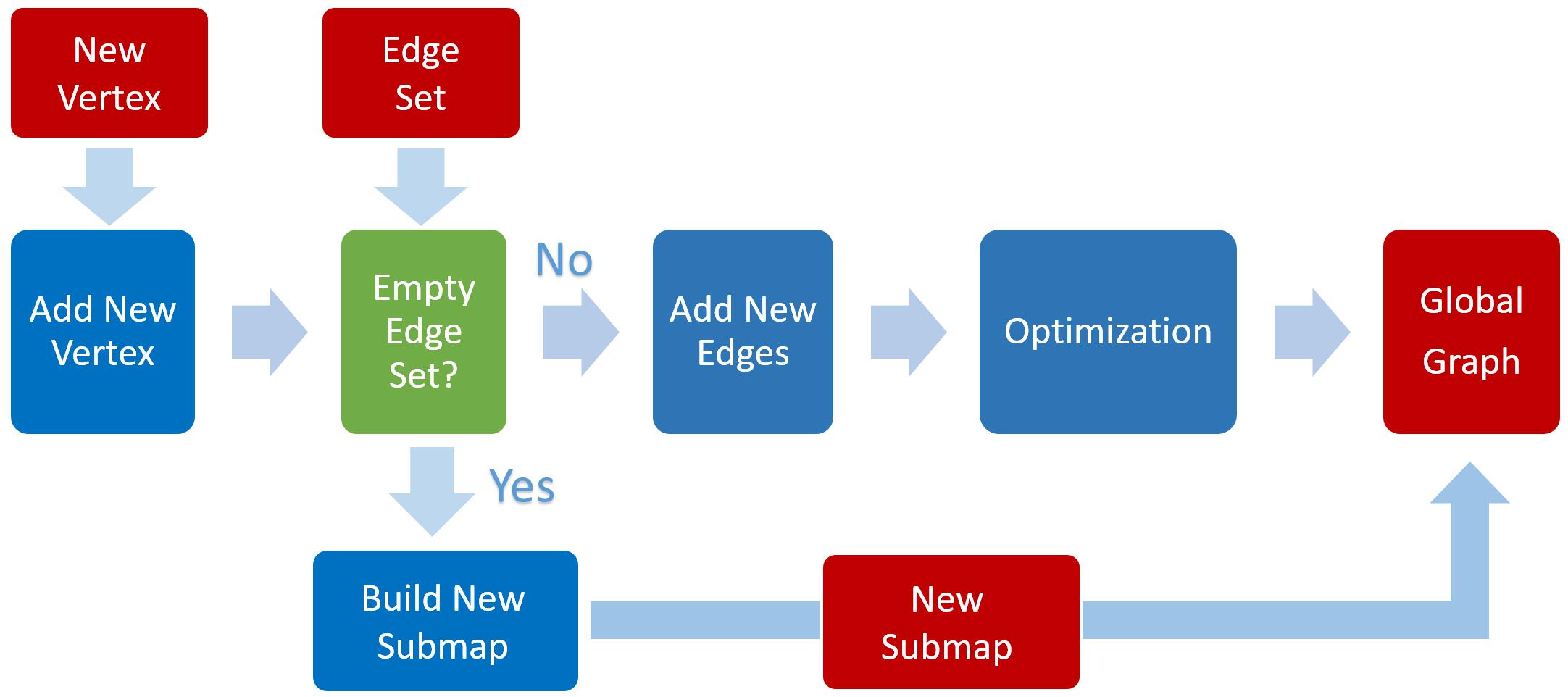}
\caption{Back-end pipeline}
\label{fig_back_end}
\end{figure}

In the front-end, when a frame is lost, which means its edge set is empty, we take this frame as the first keyframe of a new submap and try to maintain tracking in this submap. The first vertex of a new submap is placed at the origin and it is not fixed. This approach makes the system update the graph all the time: maintaining tracking in the current submap or building a new submap. With respect to the back-end, a new submap is equal to an independent set of vertices and edges, as is shown in Fig. \ref{fig_submap}. The back-end containing multiple submaps is called submap-based back-end, the pipeline of the approach is illustrated in Fig. \ref{fig_back_end}.

Within a submap, an estimation is given to the vertex using a motion model. The estimation is calculated from the relative pose and the last keyframe pose tracked in the same submap. We always try to track the frame in the current submap, and the tracking calculation is a two views geometry. With such a tracking strategy, despite lost frames, our system is able to work without the information of the existing graph.

The weighted information matrix of an edge is set according to the number of inlier feature correspondences $s$ and the total number of detected features, the information matrix is set by the formula \ref{for_information_matrix},
\begin{equation}\label{for_information_matrix}
  \Xi {\rm{ = }}\left[ {\begin{array}{*{20}{c}}
1&{}&{}&{}&{}&{}\\
{}&1&{}&{}&{}&{}\\
{}&{}&1&{}&{}&{}\\
{}&{}&{}&\Omega &{}&{}\\
{}&{}&{}&{}&\Omega &{}\\
{}&{}&{}&{}&{}&\Omega
\end{array}} \right]
\end{equation}
where $\Omega {\rm{ = }}{s \mathord{\left/ {\vphantom {s n}} \right. \kern-\nulldelimiterspace} n} \bullet 100$, and $n$ is the total number of detected features. Such an information matrix makes the intensity of the translation terms in a ``spring model'' weak.

For submap merging, we build loop-closure edges between the vertices in the different submap. The loop-closure edges align the submaps after optimization. However, submap merging by place recognition only is fragile. Therefore, the switchable factors are added to the loop-closure edges. By a lax condition, many loop-closure edges are added, even though the correctness is not guaranteed. However, with the switchable factors, the loop-closure edges that cannot fit the global consistency well are closed, such a strategy is similar to the reference \cite{c15}, which adds vertices laxly and executes a ``culling'' to remove the outliers. The vertices that have no edge or whose edges are all closed are discarded.

However, even after vertex discarding, a large number of vertices and edges are inserted after a long time running, which increases the optimization time consumption. Hence, an exploration end-condition is designed to stop the exploration and limit the growth of the graph, which will be introduced in the next section.

\section{Implementation}
\label{section_implementation}

For a complete exploration system, beside the methods mentioned above, other implementation details are introduced in this section. Also, to prevent an infinite exploration in a limited space, an end-condition is proposed to stop the exploration, which depends on the density of vertices. As for system building, to enhance the reliability of our system, we applied some mature open sources to our implementation, such as G2O \cite{c18}, OpenCV \cite{c19} and Eigen \cite{c20}, which are reliable.

\subsection{Exploration end-condition}

With respect to the growth of the graph, the number of edges and vertices increases during a dense exploration. However, no marginalization is performed in our back-end, to avoid the risk of breaking submap merging. In addition, a complete environment representation is beneficial to robust visual localization. Hence, an end-condition is needed to evaluate the completeness of environment mapping and limit the growth of graph.

\begin{figure}[htp]
\centering
\includegraphics[width=0.35\textwidth]{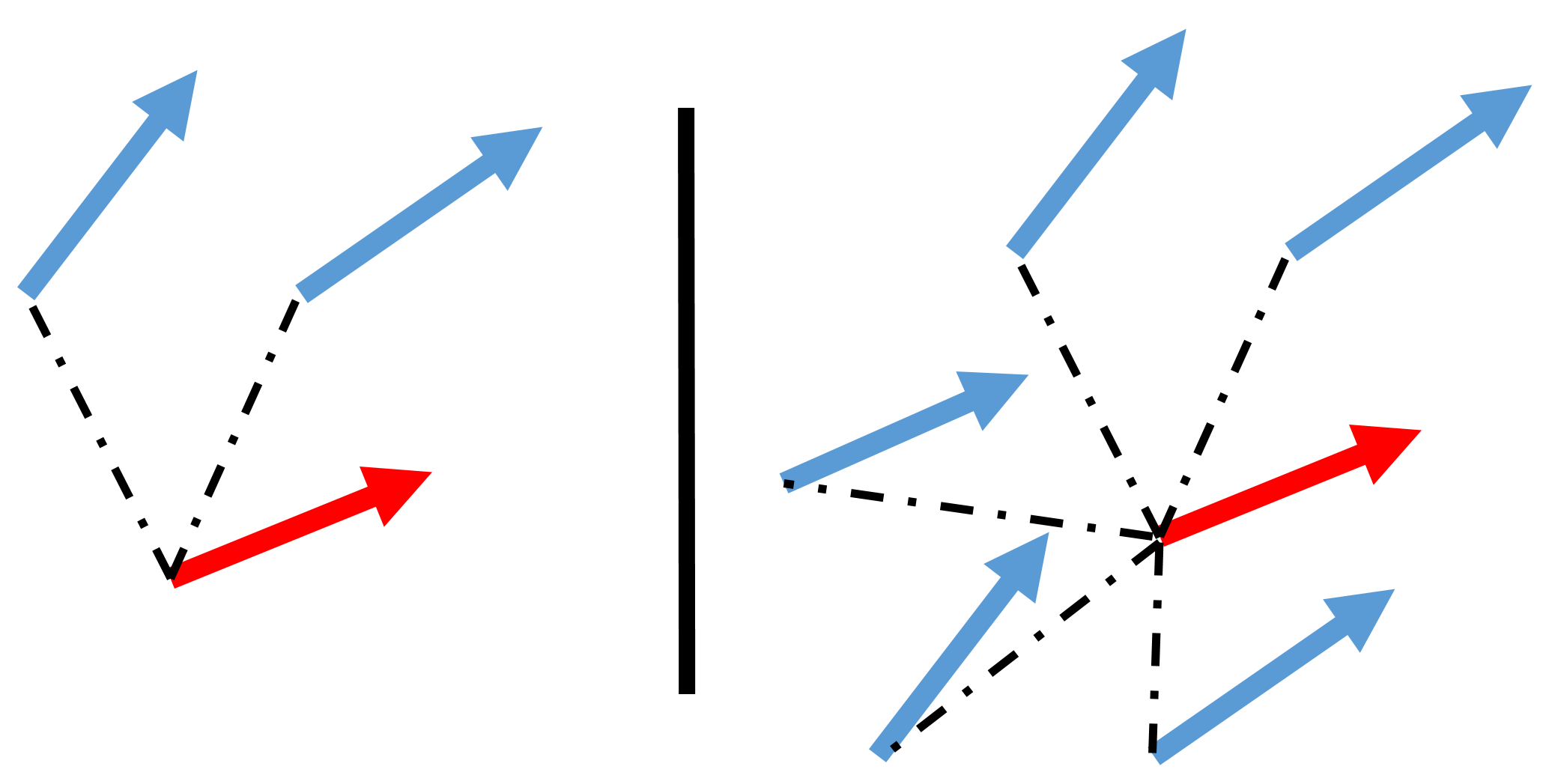}
\caption{Illustration of the end-condition, red arrow is the new vertex, blue arrows are the existing vertices, and the black dotted lines are edges. The left figure: when the existing vertices are not dense, less edges can be established between the vertices. Right figure: when the existing vertices become dense, more edges can be established.}
\label{fig_ending_condition}
\end{figure}

Considering a dense exploration, the vertices in a limited space become denser and denser. Therefore, the exploration can be stopped if the vertices are dense enough. Since the number of loop-closure edges is proportional to the overlap between the keyframe observations, and the overlap also increase with the increasing of vertex density. In other words, when the number of edges (including tracking edges and loop-closure edges) for a new vertex are big enough, and such a situation continues for some time, the vertices in a limited space can be thought to be dense enough. This end-condition is illustrated in Fig. \ref{fig_ending_condition}.

To implement such an end-condition, a threshold in terms of the size of new edges is set, represented as $T_e$. The edges and vertices added to the graph after the last optimization are called new edges and new vertices, the vertices left after vertex discarding are called active vertices. When the size of new edges is greater than $T_e$, an optimization is performed and the size of new active vertices $S_{nv}$ is recorded. If $S_{nv}$ is not greater than $T_e/(k+1)$ (which means all the potential loop-closures are accepted and the tracking edge is established), a counter $c$ plus one. Only when $c$ is large enough, we stop the exploration and shutdown the system.

\subsection{Relative pose calculation implementation}

As is mentioned in Section \ref{section_front_end}, a VO is applied to relative pose calculation. During feature matching, the minimum $L1$ distance $m_d$ of the correspondences is recorded, and only the correspondences with a distance in the range of $p \cdot {m_d}$ are considered, where $p$ is given. For the fundamental matrix finding, we refine the matrix with the RANSAC inlier correspondences after the first matrix finding.

\subsection{Optimization implementation}

After building a graph, two types of optimization are found in the system: ``free-time optimization'' and ``regular optimization''. Since not all the frames are regarded as keyframes as mentioned in Section \ref{section_front_end}, we optimize the graph with a bigger iteration time when the front-end is free, and such an optimization is called ``free-time optimization''.

Also, to meet the end-condition mentioned in Section \ref{section_back_end}, an optimization triggered by $T_e$ is performed and called ``regular optimization'', which has a smaller iteration time. Such an optimization is required since some vertices may become inactive when switchable factors closing during optimization, and the size of active vertices is required by the end-condition.

\section{Experiments}
\label{section_experiments}

For evaluating the precision and robustness of our system, two challenging datasets are collected, which contain the image sequence of dense exploration in our lab captured by a cheap monocular camera. Also, the GT (Ground Truth) trajectory collected from the Opti-Track motion capture system is recorded for the precision analysis with ATE RMSE \cite{c22}. In addition, it is hard to find the existing datasets meeting our requirements, which should be a dense exploration process and the duration should be long enough to meet the end-condition.

As a popular VSLAM system, ORB-SLAM 2.0 (abbreviated as ORB-SLAM in this section) is selected for comparison, which is also an ORB feature-based VSLAM system.

Also, another experiment is designed for relocalization evaluation. We run a part of the dataset for graph building and keep the graph fixed after meeting the end-condition. Then we run another part of the dataset for relocalization evaluation with that fixed graph. Compared with the first experiment, the input of the relocalization evaluation is a set of independent images instead of an image sequence. The experiment is designed to demonstrate the ability of solving the ``kidnapped'' problem after dense exploration with our system.

In the end of this section, we analyse the experimental results and discuss the system.
\subsection{Evaluation and Comparison}

\begin{figure}[htp]
\centering
\includegraphics[width=0.4\textwidth]{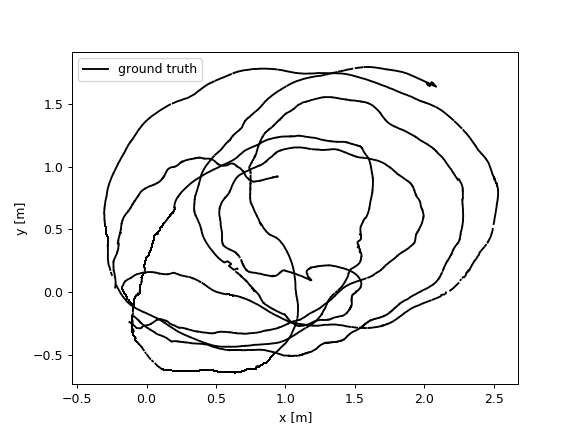}
\caption{Ground truth trajectory of dataset 1}
\label{fig_dataset_1}
\end{figure}

\begin{figure}[htp]
\centering
\includegraphics[width=0.4\textwidth]{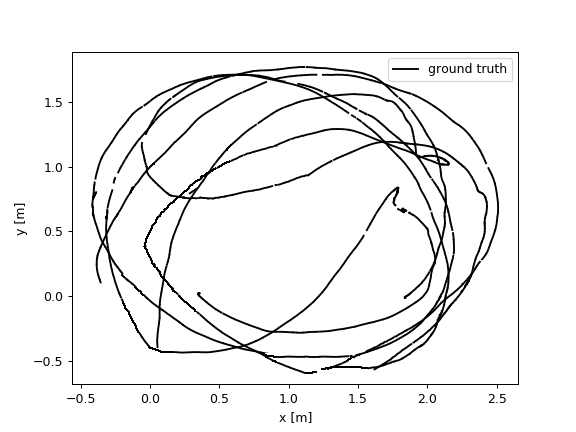}
\caption{Ground truth trajectory of dataset 2}
\label{fig_dataset_2}
\end{figure}

Two dense exploration datasets are collected, the GT trajectories are shown in Fig. \ref{fig_dataset_1} and Fig. \ref{fig_dataset_2}. We explore a limited space repeatedly and observe the scene in different perspectives during datasets collection. The durations of the datasets are long enough to meet the end-condition.

\begin{figure*}[htp]
\centering
\includegraphics[width=0.60\textwidth]{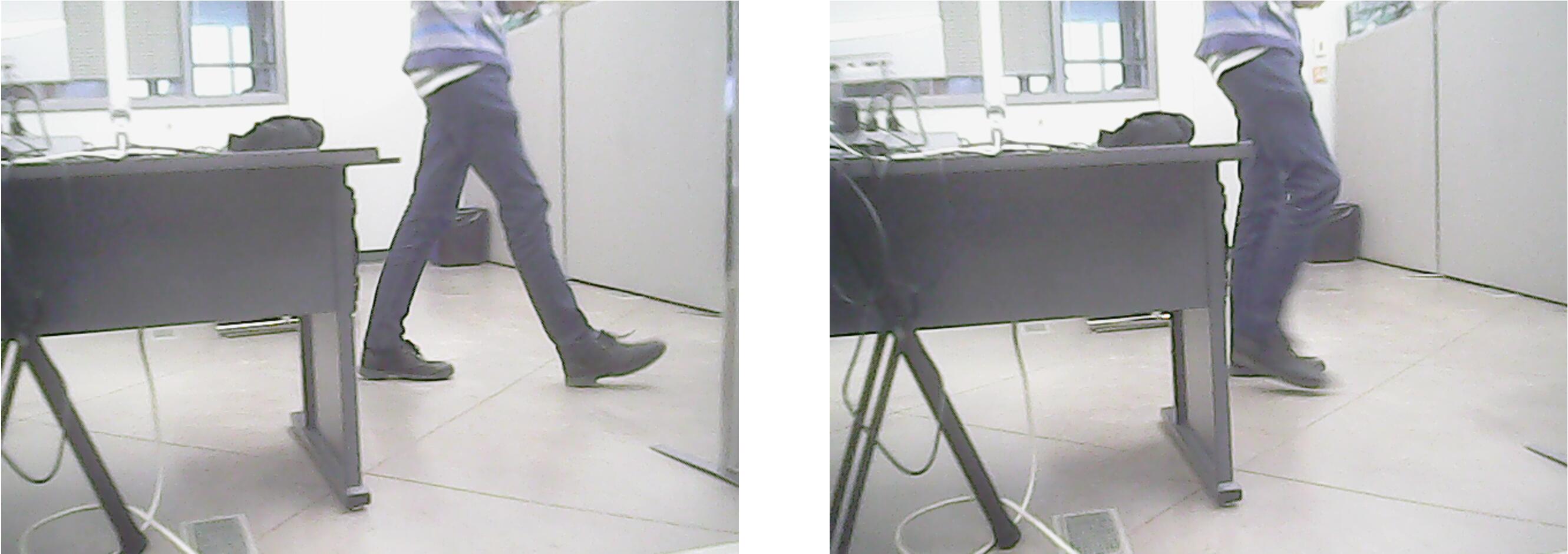}
\caption{Dynamic objects}
\label{fig_dynamice_object}
\end{figure*}

\begin{figure*}[htp]
\centering
\includegraphics[width=0.75\textwidth]{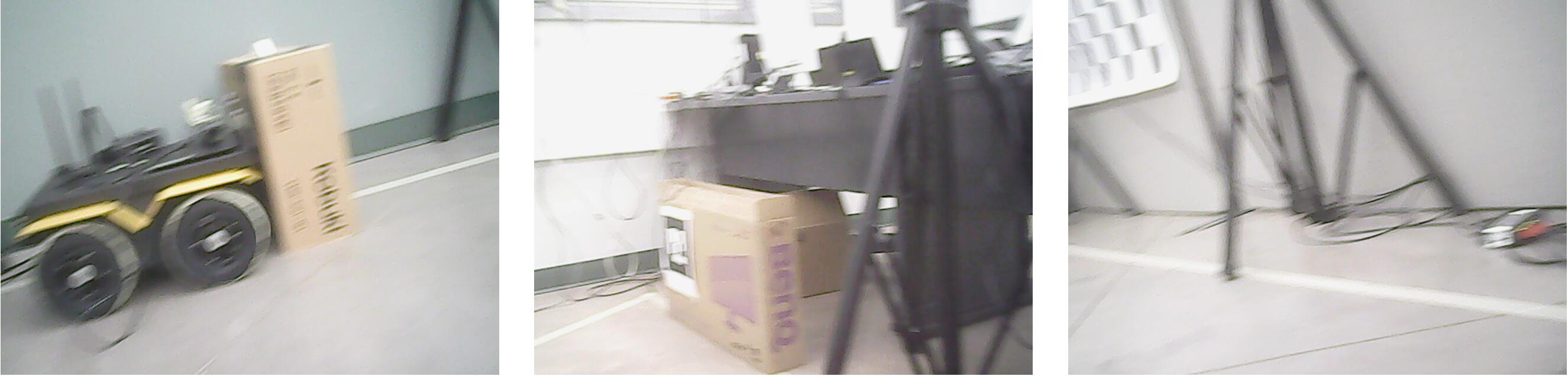}
\caption{Motion blurred frames}
\label{fig_motion_blur}
\end{figure*}

\begin{figure*}[htp]
\centering
\includegraphics[width=0.75\textwidth]{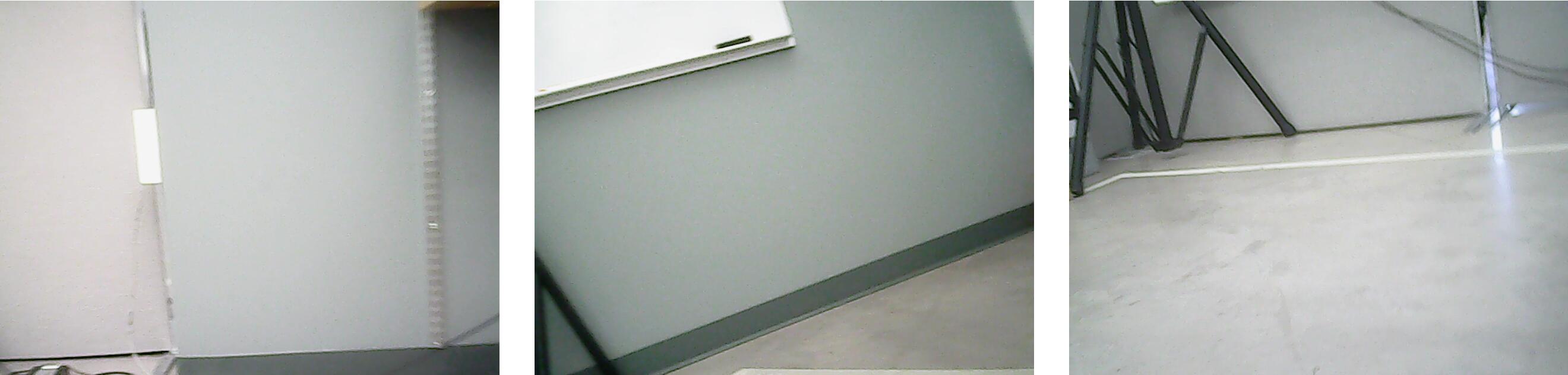}
\caption{Textureless walls and floors}
\label{fig_Textureless}
\end{figure*}

Some challenging situations in our collected datasets are shown in Fig. \ref{fig_dynamice_object}, Fig. \ref{fig_motion_blur} and Fig. \ref{fig_Textureless}, including the dynamic objects, textureless walls and floors, and motion blurred frames, which are typically challenging situations for VSLAM.

Since the RMSE performances of those challenging datasets are unstable, we run 6 times for one dataset to compare our system with ORB-SLAM. In addition, we find that the maximum number of detected features in ORB-SLAM influences its performance, therefore, two parameters are set and experimented: 1000 and 1500. However, the maximum number of detected features in our system is always 500. For a fair evaluation, the duration for meeting the end-condition is recorded, and ORB-SLAM is run for the same duration.

Beside the RMSE, the tracking percentage (TP) of ORB-SLAM is indicated, which equals the tracked frames of ORB-SLAM divided by the frames of the whole dataset. The precision comparison and the TP are shown in Tab. \ref{tab_precision1} and Tab. \ref{tab_precision2}, where the second and the third columns are the RMSE of our system and ORB-SLAM, the ``Feature Number'' is the feature detection parameter in ORB-SLAM. The results of our system and ORB-SLAM in the same row are run for the same duration.

\begin{table}[h]
\centering
\begin{tabular}{|c|c|c|c|c|}
\hline
                & Our system & ORB-SLAM  & Feature Number & TP \\
\hline
Test 1          &    0.615   &   0.884   &   1500   &   31.0\%  \\
\hline
Test 2          &    0.602   &   0.668   &   1500   &   21.7\%  \\
\hline
Test 3          &    0.696   &   0.922   &   1500   &   78.4\%  \\
\hline
Test 4          &    0.737   &   0.921   &   1500   &   71.5\%  \\
\hline
Test 5          &    0.593   &   0.412   &   1000   &   10.0\%  \\
\hline
Test 6          &    0.697   &   0.558   &   1000   &   18.6\%  \\
\hline
\end{tabular}
\caption{Precision comparison for dataset 1}
\label{tab_precision1}
\end{table}

\begin{table}[h]
\centering
\begin{tabular}{|c|c|c|c|c|}
\hline
                & Our system &  ORB-SLAM & Feature Number & TP \\
\hline
Test 1          &    0.457   &   1.000   &   1500   &   63.2\%  \\
\hline
Test 2          &    0.468   &   1.012   &   1500   &   48.0\%  \\
\hline
Test 3          &    0.432   &   0.925   &   1500   &   46.3\%  \\
\hline
Test 4          &    0.341   &   0.728   &   1500   &   26.5\%  \\
\hline
Test 5          &    0.408   &   0.676   &   1000   &   19.0\%  \\
\hline
Test 6          &    0.503   &   0.863   &   1000   &   18.6\%  \\
\hline
\end{tabular}
\caption{Precision comparison for dataset 2}
\label{tab_precision2}
\end{table}

\begin{figure}[htp]
\centering
\includegraphics[width=0.4\textwidth]{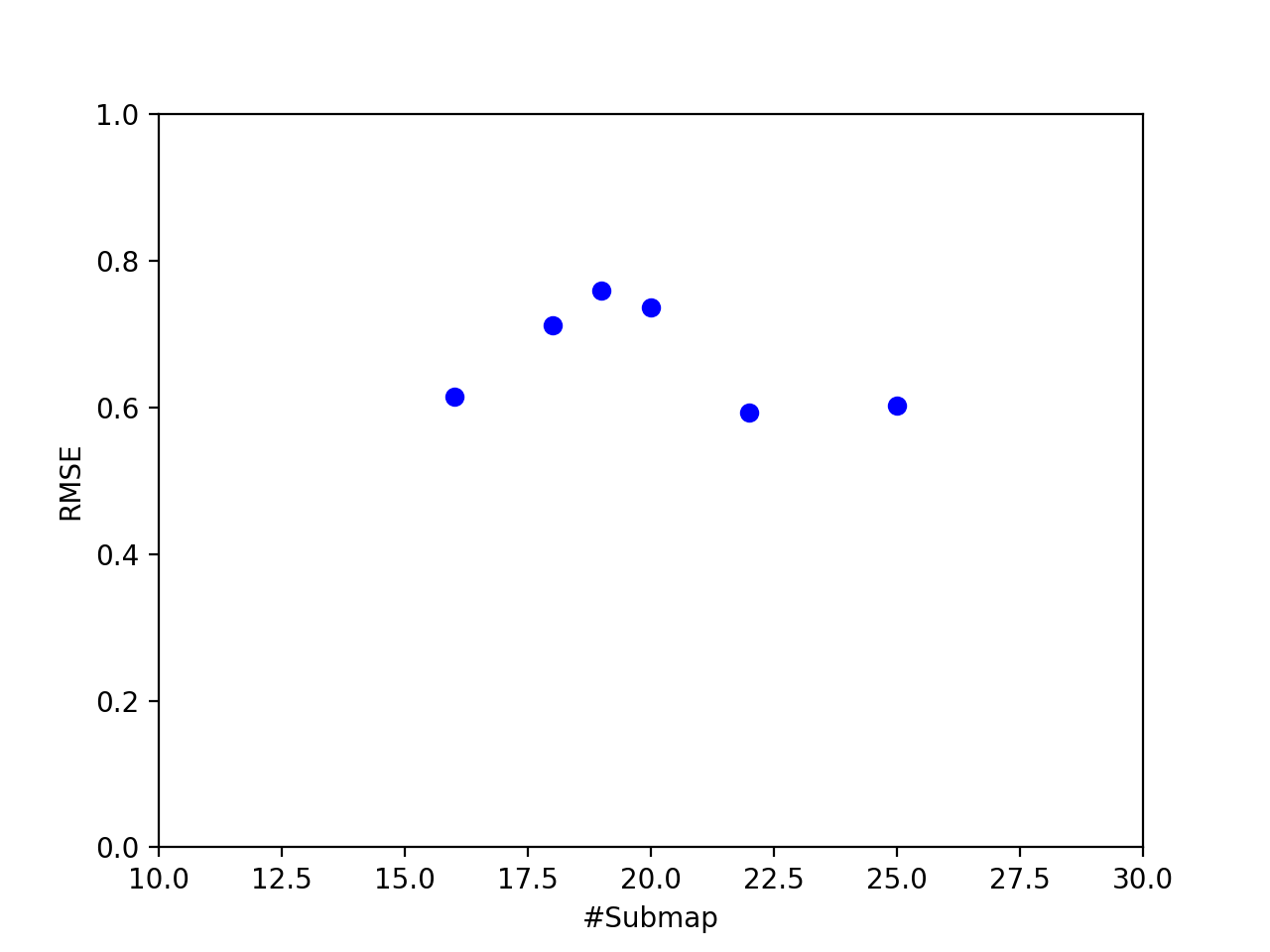}
\caption{Scatter plot of the number of submaps and RMSE for the first dataset}
\label{fig_submap_RMSE1}
\end{figure}

\begin{figure}[htp]
\centering
\includegraphics[width=0.4\textwidth]{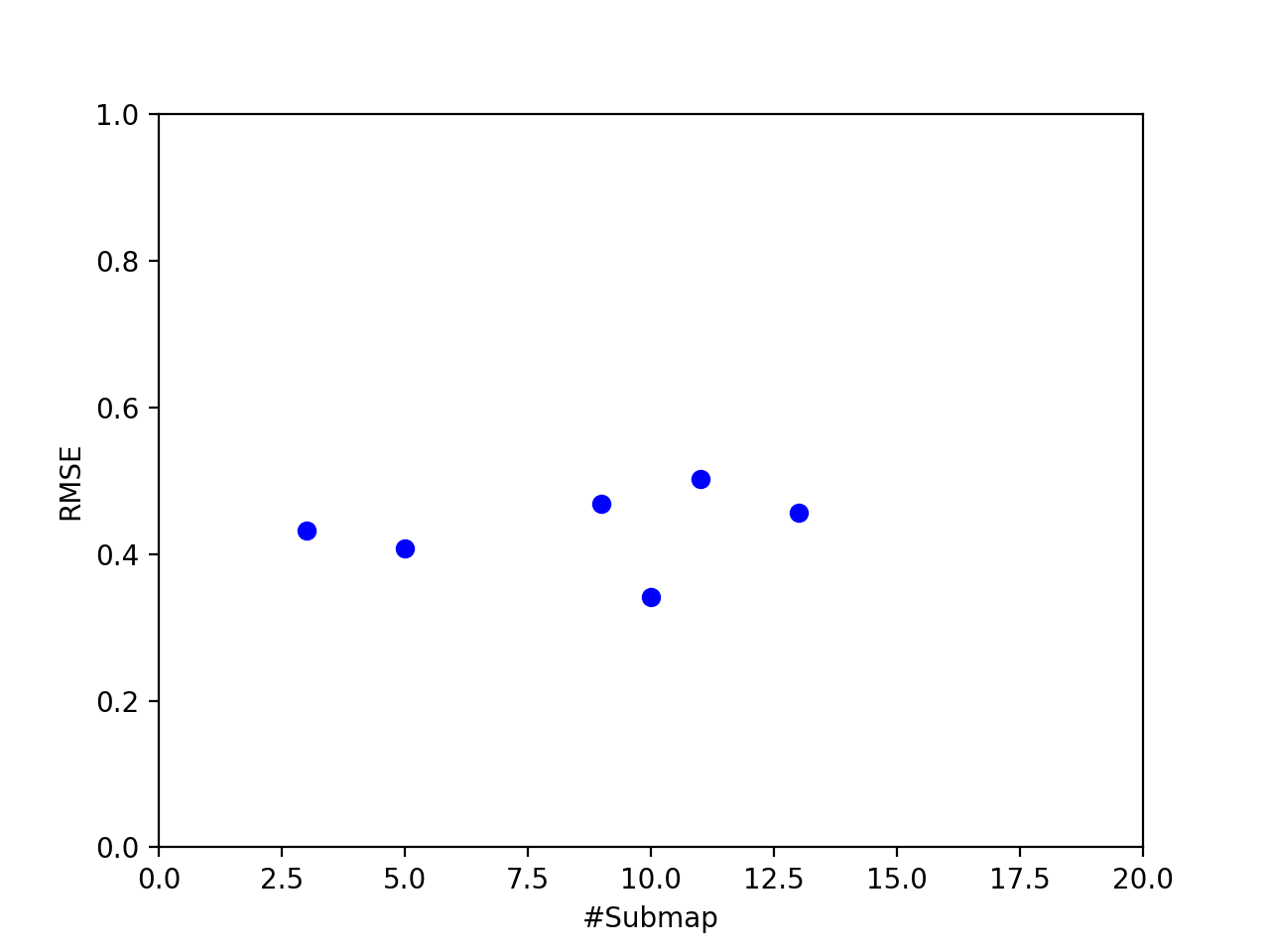}
\caption{Scatter plot of the number of submaps and RMSE for the second dataset}
\label{fig_submap_RMSE2}
\end{figure}

Since our system updates the graph all the time, the TP in our system is always 100\%, it is not indicated in the tables. Instead, the number of submaps, which is proportional to the number of tracking lost, is recorded. The scatter plots with respect to the number of submaps and the RMSE are shown in Fig. \ref{fig_submap_RMSE1} and Fig. \ref{fig_submap_RMSE2}. The correlation coefficients of the samples drawn in Fig. \ref{fig_submap_RMSE1} and Fig. \ref{fig_submap_RMSE2} are -0.369 and 0.240, which mean the correlation between the size of submaps and the RMSE is not significant.

In addition, we pick the tests with the same number of submaps to evaluate the stability of our system. The average of the RMSE among the samples is 0.513, and the standard deviation is 0.046.

To compare the robustness of ORB-SLAM and our system, beside the TP, we record the number of new keyframes in ORB-SLAM (with 1500 features) before each lost tracking, and calculate the average (A-KFs) and the standard deviation (SD-KFs) of that number. Also, the same metrics are recorded for our system. Beside the A-KFs and SD-KFs, the average RMSE (A-RMSE) of the tests (6 pairs of tests are run for each dataset) is shown in the Tab. \ref{tab_robustness1} and Tab. \ref{tab_robustness2}. Even though our system updates the graph all the time, because of the keyframe selection conditions and the difference between the frame rate of our system and the datasets, not all the frames are considered.

\begin{table}[h]
\centering
\begin{tabular}{|c|c|c|c|}
\hline
                & A-RMSE &  A-KFs & SD-KFs \\
\hline
ORB-SLAM        & 0.708 &  13.88 &   2.37 \\
\hline
Our system      & 0.692 &  1069.20 &  23.60 \\
\hline
\end{tabular}
\caption{Robustness comparison for dataset 1}
\label{tab_robustness1}
\end{table}

\begin{table}[h]
\centering
\begin{tabular}{|c|c|c|c|}
\hline
                & A-RMSE &  A-KFs & SD-KFs \\
\hline
ORB-SLAM        & 0.893 &  33.54 &   3.04 \\
\hline
Our system      & 0.421 &  836.80 &  20.80 \\
\hline
\end{tabular}
\caption{Robustness comparison for dataset 2}
\label{tab_robustness2}
\end{table}
\subsection{Relocalization}
We divide our collected dataset into two parts for graph building and relocalization evaluation. 80\% of the images are used for graph building and 20\% for ``kidnapped'' relocalization evaluation. These two processes are run in order.

In the relocalization evaluation, the parameter $k$ for place recognition is set higher than the first experiment for adding more edges, and other parameters are kept the same. All the vertices after graph building are set to be fixed, and the vertices that have been optimized are also set to be fixed. In addition, no tracking edge is established in this experiment, since we try to solve a ``kidnapped'' relocalization problem whose observation is an independent image instead of an image sequence.

Two performance metrics are computed: relocalization precision and tracking percentage. To avoid missing frames during the evaluation, we play the relocalization dataset in a low frame rate and remove the keyframe selection conditions, in order to be able to consider more frames.

\begin{figure}[htp]
\centering
\includegraphics[width=0.45\textwidth]{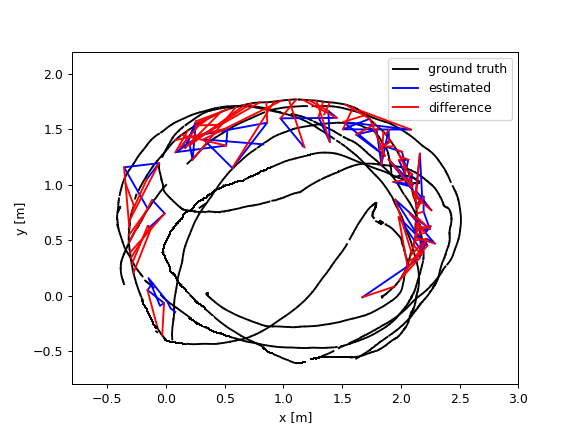}
\caption{Result of relocalization and the GT. The red lines are the difference between the estimated poses and the correspondent GT poses. The shorter the line, the smaller the error.}
\label{fig_relocalization2}
\end{figure}

The precision is also calculated in terms of ATE RMSE, and the tracking percentage equals the tracked frames divided by the considered frames. The experimental results are shown in Fig. \ref{fig_relocalization2}, whose tracking percentage is 65.8\% and the RMSE is 0.312 based on a fixed graph with a 0.308 RMSE.

For ORB-SLAM, since no complete trajectory can be built in our collected dataset, no relocalization evaluation is performed.

\subsection{Discussion}

As shown in Tab. \ref{tab_precision1} and Tab. \ref{tab_precision2}, in most cases, our system has a smaller RMSE in both datasets, which is achieved by the design of our system. Having more edges leads to better precision after optimization, although the original intention of the multi-constraint front-end is improving the robustness. Also, with a high optimization efficiency resulting from the pose-graph, more optimization iterations can be run. ORB-SLAM has a smaller RMSE than our system in Test 5 and Test 6 in the first dataset, which could be due to the low TP (10\% and 18.6\%) with a small number of samples in RMSE calculation.

Since the number of submaps is proportional to the number of tracking failure, as shown in Fig. \ref{fig_submap_RMSE1} and Fig. \ref{fig_submap_RMSE2}, due to the submap-based back-end, our system keeps working even if no edge can be established for that vertex. As the correlation coefficients and Figs. \ref{fig_submap_RMSE1} and \ref{fig_submap_RMSE2} shows, the RMSE does not increase with the increasing of the size of submaps, verifying the ability to overcome tracking failure and the robustness of our system.

As is shown in Tab. \ref{tab_robustness1} and Tab. \ref{tab_robustness2}, the A-KFs of ORB-SLAM is much smaller than our system, since our system keeps building submaps and updates the global graph all the time. Also, in terms of precision (A-RMSE), our system is also better.

For some challenging situations, instead of trying to maintain tracking with those frames, we just stop building the current submap and initialize a new submap (even some submaps are discarded since no frame can be tracked). In our opinion, lost tracking is a sequential perspective problem, and the submaps can be tracked in other perspective in another sequence.

Regarding relocalization, our system obtains a 0.312 ATE RMSE, as well as a 65.8\% tracking. Thanks to the proposed end-condition, we are able to create a topologically dense map (dense keyframes and complete images in each keyframe) after exploration. With such a map, even an observation in a different perspective can be relocalized. We believe that such a dense map is practical for visual navigation, and that the dense exploration is possible in most visual exploration applications.

As for the time consumption, the average frame rate of our system in the experiment is 3.0 frames per second (the frame
rate while collecting our datasets is 30), which is lower than the frame rate of ORB-SLAM. Such a low frame rate is caused by loop-closure detection and verification for each frame, both taking much time. However, thanks to our feature-based front-end, whose baseline is allowed to be wide, tracking is able to continue to operate. Even if a frame cannot be tracked from the last keyframe, because of the multiple constraints strategy, the system can still calculate the frame poses.

With the experimental results and the analysis above, we have provided evidence for our contributions: a robust VSLAM system with a submap-based back-end, whose precision is also improved by a multi-constraint front-end. The map whose completeness is evaluated by the proposed end-condition has the ability to localize the frames in different perspectives.

\section{Conclusion}
\label{section_conclusion}

A VSLAM system is proposed in this paper. It is designed for robust exploration and localization in a limited space. A submap-based back-end is introduced to improve the robustness. Also, a front-end is designed with place recognition, building multiple constraints for a frame and adding more edges to the graph to improve the robustness as well as the precision. Some challenging datasets are collected to evaluate our system, and our system has a smaller ATE RMSE and a more robust performance than the state-of-the-art. Also, the ability to solve ``kidnapped'' problem is demonstrated.

In the future, we will design a complete autonomous exploration system based on this study. The place recognition module and the relative pose calculation module can be improved by other alternative approaches, such as CNN based image matching and deep-feature based tracking, to further improve the performance of the VSLAM system.

\begin{appendices}
\end{appendices}

\end{document}